# *AutoPureData*: Automated Filtering of Undesirable Web Data to Update LLM Knowledge

**Praneeth Vadlapati**
*University of Arizona*
praneethv@arizona.edu
ORCID: 0009-0006-2592-2564

*Abstract*—Up-to-date and reliable language models are consistently sought after and are essential in various applications. Typically, models are trained on a fixed dataset and then deployed globally. However, the knowledge of the models becomes outdated. Enabling automatic updation of AI knowledge using web data involves significant concerns regarding the model's safety and quality due to a threat from unsafe and undesirable text across the web. The purity of new data was essential for updating knowledge of language models to maintain their reliability. This paper proposes AutoPureData, a system that automatically collects and purifies web data. The system loaded a sample of web data. Utilizing existing trusted AI models, it successfully eliminated unsafe text with an accuracy of 97% and undesirable text with an accuracy of 86%, demonstrating the system's effectiveness in purifying the data. The system ensures that only meaningful and safe text can be used to update LLM knowledge. The pure text was then optimized and stored in a vector database for future querying. It was found that LLM can fetch new data from the vector DB. The LLM writes the RAG query in English, even if the user's query is in another language, proving that the system can perform cross-lingual retrieval. This paper proposes a method to maintain the accuracy and relevance of up-to-date language models by ensuring that only purified data was used to update LLM knowledge. This work contributes to updating knowledge of chatbots using meaningful and safe text, enhancing their utility across various industries, and potentially reducing the risks associated with outputs caused by unsafe or impure data. Code is available at https://github.com/Pro-GenAI/AutoPureData.

## I. INTRODUCTION

The web is a vast source of information [1]. However, the reliability and quality of the information vary significantly [2]. "Garbage in, garbage out" indicates that the input data used for training or fine-tuning a Language Model impacts the quality of the resultant model [3], [4], [5]. Data quality is crucial to updating model knowledge, as using unsafe or impure data would compromise the model quality. Using search engines on demand is often time-consuming and expensive, and web data is reliable only after purification. Organizations automate the data collection process, yet not the filtration process [6].

### A. Challenges with Manual-Only Data Filtering

Human reviewers are employed to play a crucial role in manually maintaining data quality. However, relying on manual-only data filtering introduces human bias and errors [7], necessitating review by multiple reviewers to avoid biases and mistakes. This long process causes a delay in the data preparation process, preventing models from staying up-to-date, especially when new data is constantly created in multiple languages. A considerable amount of text remains to be unexplored with manual review [8]. Additionally, some detections, such as hidden biases, are not made by humans without AI assistance. Hence, it is essential to filter out undesirable text in an automated manner.

### B. Proposed Solution and Its Benefits

This paper proposes a system for an automated filtration of web data using existing trusted AI models, followed by the usage of new filtered data to update the knowledge of Large Language Models (LLMs). Natural Language Processing (NLP) tasks can be performed using existing trusted LLMs [9], [10]. The new system aims to ensure a high quality of data, which is crucial for the success of AI models. Additionally, the system was designed to filter data from "untrusted sources", even if the text appears safe, further enhancing the reliability of the filtered data. Some attacks on AI models, such as Data Poisoning attacks, could be avoided by processing the training data [11].

New filtered data was stored in a Vector Database (DB) and accessed using Retrieval-Augmented Generation (RAG) with system-prompting to generate responses to user queries. According to research, RAG with system-prompting was more effective in utilizing new data when compared to fine-tuning [12], [13]. The quality of up-to-date LLMs helps organizations retain users and prepare for future regulatory requirements to save a substantial loss caused by unsafe or low-quality models. The proposed system significantly reduces the time and effort required for data filtration, thereby increasing the efficiency of the data preparation process, which is a crucial part of knowledge updation.

## II. LITERATURE REVIEW

Li et al. (2023) [1] explored the usage of search engines to fetch the latest data to answer queries but did not focus on detecting undesirable text that impacts model responses. Penedo et al. (2024) presented the FineWeb dataset [14] with refined and deduplicated web data suitable for training LLMs but did not focus on detecting undesirable text. Yexiao He et al. (2024) [15] introduced SHED, a method for Automated Dataset Refinement that selects the most informative data for training. Biester et al. (2024) [16] introduced LLMClean, which includes automated data cleaning using rule-based and ML-based cleaning tools. Similarly, Chen and Mueller (2023) [17] worked on automated data curation of data for fine-tuning. In existing work, the priority was on creating usable datasets without a focus on the removal of undesired data from diverse data sources such as the web. This paper presents an automated filtering system based on an analysis of the gap in existing research.

### III. METHODS

#### A. Data Collection

The system collected 100 rows of web data from the FineWeb dataset, known for refined and deduplicated content [14], [18]. The web data was diverse and originated from various web pages, ensuring the system was tested across different contexts.

TABLE I. SAMPLE DATA

| ID | URL | Text |
|---|---|---|
| a1 | http://38.paulosimoes.net/ | We want to know how to best serve you. Please use one of the forms below… |
| a2 | http://aberdeencreekfl.com/ | Architectural Control Committee Policies and Forms… |

#### B. Data Flagging

Considering multiple ways to detect undesirable text in web data, the system uses a multi-step flagging process using existing trusted AI models.

*1) Flagging Unsafe Text and Domains:*

Unsafe text and domains are flagged using LLaMaGuard 2 [19]. According to the Model Card page, LLaMaGuard 2 has an F-1 score of 91.5% and a False Positive Rate of 4% and was noted to be superior to other popular moderation models or APIs [19]. Domains are extracted from URLs.

*2) Flagging Unreliable Domains:*

A search engine was utilized to determine whether a domain was indexed. Non-indexed domains are flagged in this step based on the assumption that search engines do not index unreliable domains.

*3) Flagging Undesirable Text Using an LLM:*

LLaMa 3 (70B) [20] was the language model used to detect other undesirable text using a set of rules and a list of target flags provided as input, such as unusable (non-informative) content, advertisements, sensitive topics, biased information, and other undesired content like religious content, lottery, scam, and data poisoning attempts.

Prompt Template 1. **Prompt to flag undesirable text using an LLM**

> You are a content moderator. The text below will be used to fine-tune LLMs. Fill the 'flags' column with one or more flags to detect from: `{flags_to_detect}`.
> If you flag a row, fill the 'flag_reason' column with a very short reason for flag choices. Return back only CSV text in triple backticks and no other text, like
> ```id, flags, flag_reason
> a1, "safe", "No flags"
> a2, "scam,spam", "Suggests a potential crime"```
>
> Input data: ```{csv_text}```
> Output columns: id, flags, flag_reason

Prompt Template 2. **Prompt to flag non-informative text using an LLM**

> You are a content moderator. The text below will be used to fine-tune LLMs. Use "unusable" as the flag if the text does not convey new information, or else mark it as "safe". Return only CSV text in triple backticks and no other text, like
> ```id, unusable_flag, unusable_flag_reason
> a1, "unusable", "No useful/new information"
> a2, "safe", "Useful information"```
>
> Input data: ```{csv_text}```
> Output columns: id, unusable_flag, unusable_flag_reason

#### C. Human Expert Review and Further Processing

The flagged data underwent a human review to correct the flags where necessary. The review allows us to calculate the accuracy of AI-based flags. After the review, Flagged rows were removed from the dataset to ensure data purity. An LLM was used to optimize the text by making it concise and enhancing the efficiency of the LLM responses after integrating new data.

Prompt Template 3. **Prompt to shorten and optimize the text before usage**

> You are a content moderator preparing a dataset for fine-tuning a language model. You have a text that needs to be shortened and made suitable for fine-tuning. Retain important details like date and location. Return the optimized text in triple backticks.
>
> Original text: ```{original_text}```

#### D. Integrating new data with the LLM using RAG

Filtered data was stored in a Vector DB and integrated with the system using RAG. The model creates an optimized RAG query in English to enhance search efficiency, independent of the language of the user query, as the collected data was in English. As the model supports other languages, the system was considered capable of multilingual interactions with retrieval and answering using new data. The three most relevant results were retrieved from the Vector DB using the generated query and then passed to LLM as a system-prompt, providing the model with the most relevant new information.

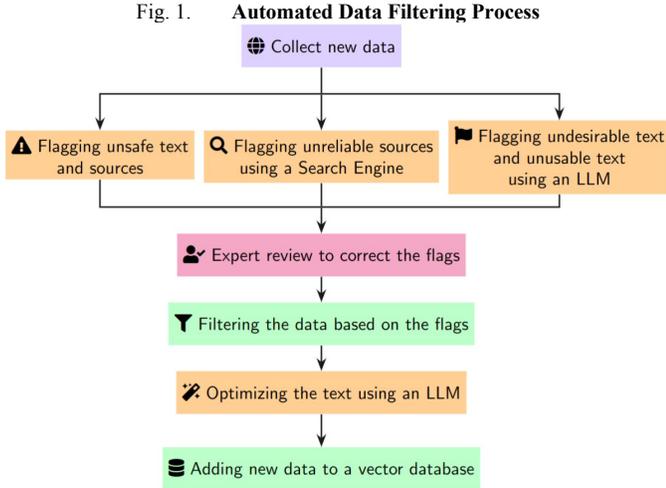

Fig. 1. **Automated Data Filtering Process**

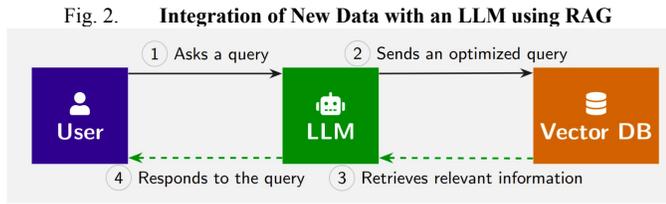

Fig. 2. **Integration of New Data with an LLM using RAG**

## IV. RESULTS

### A. Manual Correction Results

The manual correction results are presented in the table below, comparing the values finalized by human reviewers (actual values) with those predicted by AI. The values in the confusion matrices are true and false positives and negatives for the flags predicted at both stages. The LLM-generated flags, accompanied by short explanations, provided new insights, thoughts, and considerations for the human reviewer, demonstrating the usefulness and reliability of the models in the flagging process. It is important to note that the perspective of the human reviewer(s) might have an impact on the flag correction, and hence the calculation of the accuracy of the models.

TABLE II. CONFUSION MATRIX FOR "FLAGGED AS UNSAFE"

| Metric | Positive | Negative | Total |
|---|---|---|---|
| True | 7 | 90 | 97 |
| False | 1 | 2 | 3 |
| **Accuracy: 97.00%** | | | |

TABLE III. CONFUSION MATRIX FOR "FLAGGED AS UNDESIRABLE"

| Metric | Positive | Negative | Total |
|---|---|---|---|
| True | 60 | 26 | 86 |
| False | 1 | 13 | 14 |
| **Accuracy: 86.00%** | | | |

Assuming that the LLM using this data was resistant to rare undesirable text that does not get filtered automatically, a human review may not be necessary, highlighting the potential of this system.

### B. Flagging Results After Correction

The count of each flag following the reviewer's correction was presented in the figures below, noting that some rows have multiple flags. Reasons for removing the rows and the counts are presented in the table below. Dominant reasons for removal included the text being non-informative and not adding any new information to the LLM knowledge, as well as the text containing advertisements.

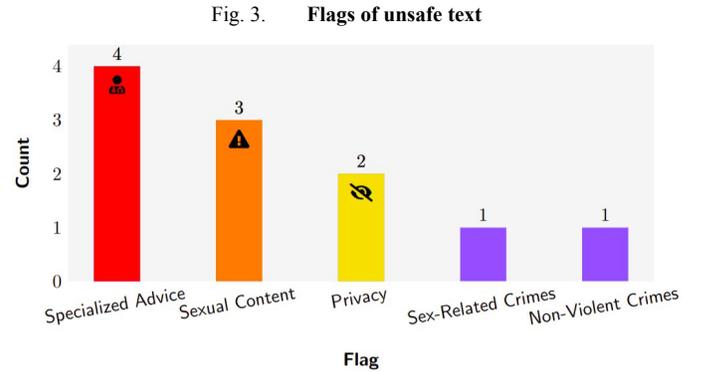

Fig. 3. **Flags of unsafe text**

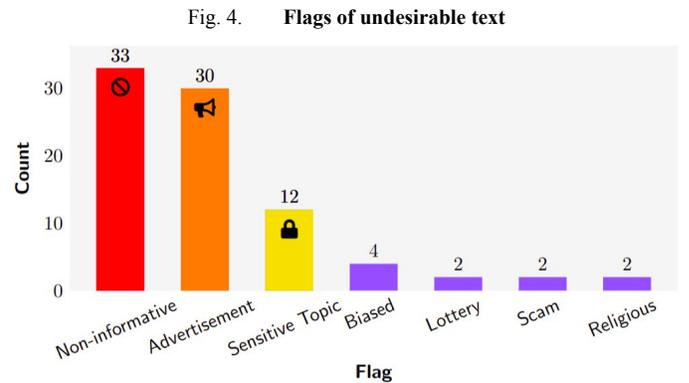

Fig. 4. **Flags of undesirable text**

TABLE IV. REASONS FOR REMOVING SOME ROWS

| Reason for removal | Count |
|---|---|
| Flagged as unsafe | 9 |
| Domain unsafe | 3 |
| Domain not indexed | 5 |
| Flagged as undesirable | 59 |
| **TOTAL** | **76** |

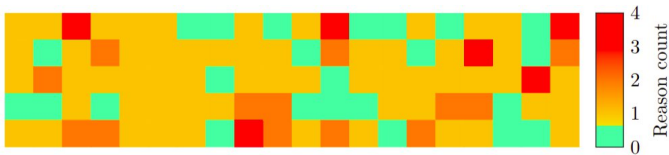

Fig. 5. Heatmap of the number of reasons/flags of each row

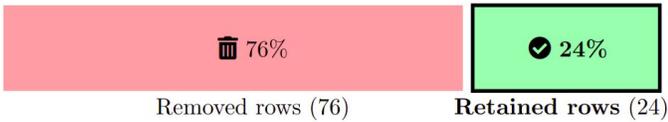

Fig. 6. Distribution of Rows: Removed vs Retained

*C. Usage in Language Models with RAG*

Integrating the system with RAG successfully allowed the utilization of newly filtered data in the models. Without RAG integration, the LLM struggled to respond to a query about new information due to a lack of new data. After integration with RAG, a noticeable improvement in response accuracy and relevance was observed. With new pure data, the LLM gained the ability to generate dependable responses to user queries. The system demonstrated the importance of updating the LLM knowledge using purified data. Below are sample responses generated by the LLM before and after integration with RAG:

❓ **Query**:

Men's Hockey Team in Bengaluru recently welcomed which new members after March 2023? On which date?

♚ **Response without retrieval**:

I'm happy to help! However, I'd like to clarify that I'm a large language model, I don't have access to real-time information or specific details about a men's hockey team.

🔍 **Response using RAG**:

The Indian Men's Hockey Team recently welcomed two new team members, Rhett Halkett (Analytical Coach) and Alan Tan (Scientific Advisor), in Bengaluru on the 6th of May 2023.

## V. DISCUSSION AND LIMITATIONS

It is essential to note that flagging does not necessitate the deployment or usage of LLMs, as alternative, faster, and more cost-effective NLP algorithms might be used. Sometimes, the flags generated by the models might require corrections by multiple human reviewers to ensure the data quality. Feedback from numerous human reviewers could be instrumental in improving the system. The system was designed to only experiment on a sample of 100 rows to test a new concept of automated filtering.

The Language Models used in this experiment are small and may not be optimal for every task. Larger and more powerful models could further improve the accuracy and reliability of the system. The system was designed to operate only in English and does not collect or process data in other languages. The experiment was focused on testing a new approach to data filtering without evaluating the system's speed, scalability, and cost-effectiveness. The data source used was only web data. Additional new data sources, such as academic journals, could be incorporated.

## VI. CONCLUSION

The system presented in the research successfully demonstrated a new capability of efficiently purifying unsafe text with an accuracy of 97% and undesirable text with an accuracy of 86%. The results mark a new step towards the development of up-to-date Language Models. The inefficiencies and potential biases in manual-only data review processes, as well as the benefits of automation in enhancing the speed, quality, and cost-effectiveness of data preparation, were explained in this paper. Organizations implementing such a system benefit from up-to-date LLMs, ultimately improving the utility of LLM-based applications while mitigating the risks associated with impure or outdated data. Organizations can benefit from saving significant time and resources, making such a system a valuable addition to their data preparation process. This research helps small organizations with limited resources to have up-to-date language models.